\documentclass[runningheads,a4paper]{llncs}

\usepackage[T1]{fontenc}
\usepackage{amsmath,amssymb}
\usepackage{algorithm}
\usepackage{algpseudocode}
\usepackage{booktabs}
\usepackage{graphicx}
\usepackage{array}
\usepackage{tabularx}
\usepackage{multirow}
\newcolumntype{C}{>{\centering\arraybackslash}X}
\usepackage[hidelinks]{hyperref}
\usepackage{url}
\usepackage{tikz}
\usetikzlibrary{arrows.meta,positioning,fit,calc}

\newcommand{\vlb}{\mathrm{VLB}}
\newcommand{\fnr}{\mathrm{FNR}}
\newcommand{\fpr}{\mathrm{FPR}}
\newcommand{\errr}{\mathrm{Err}}

\usepackage{enumitem}
\setlist{itemsep=1pt,topsep=2pt,parsep=0pt}

\let\oldthebibliography\thebibliography
\renewcommand\thebibliography[1]{%
  \oldthebibliography{#1}%
  \setlength{\itemsep}{0pt}\setlength{\parsep}{0pt}\setlength{\parskip}{0pt}}

\begin{document}

\title{Multilingual Verifier Bias in RLVR: Benchmark, Rollout Diagnosis, and the Cross-Lingual Selection Bottleneck}
\titlerunning{Multilingual Verifier Bias in RLVR}

\author{Chenyu Zhou\inst{1}\textsuperscript{*\dag} \and Qiliang Jiang\inst{2}\textsuperscript{*} \and Xu Zhou\inst{3}\textsuperscript{\dag}}
\authorrunning{C. Zhou et al.}
\institute{
School of Engineering, Institute of Science Tokyo, Japan\\
\email{zhou.c.76d6@m.isct.ac.jp}
\and
College of Control Science and Engineering, Zhejiang University, China\\
\email{jiangqiliang@zju.edu.cn}
\and
Department of Electrical and Computer Engineering, National University of Singapore, Singapore\\
\email{zhouxu\_nus@u.nus.edu}\\[0.5ex]
\textsuperscript{*}Equal contribution.
\textsuperscript{\dag}Corresponding authors.
}

\maketitle

\begin{abstract}
Reinforcement learning with verifiable rewards (RLVR) is a standard recipe for training large language models on mathematical reasoning, where an answer verifier serves as a language-neutral reward function. We show that this assumption fails in multilingual mathematical reasoning: an exact-match verifier turns format and script variation into language-dependent false-negative reward noise. We introduce a general-purpose, reusable protocol for auditing multilingual RLVR rewards: a verifier-robustness suite, a rollout-diagnosis procedure, and language-conditioned reward-error metrics for Japanese, English, and Chinese answers, applicable unchanged to any model, verifier, or language set. On MGSM EN/JP/CN rollouts with $k=8$, the exact-match proxy rejects trusted-correct answers at sharply different rates by language across Qwen3-4B, Qwen3-8B, and Llama-3.1-8B-Instruct; for Qwen3-8B, the false-negative rate reaches $0.642$ on JP against $0.122$ on EN and $0.073$ on CN, a false-negative-rate verifier language bias (VLB) of $0.569$. A same-split, plain-numeric probe localizes the mechanism to the final-answer interface: a final-answer-interface model drives the exact reward-error VLB to zero while the residual language accuracy gap is unchanged. We then expose a cross-lingual \emph{selection bottleneck}: on fresh MGSM250 rollouts, a target-local aggregation rule that uses no trusted labels closes $55\%$ and $78\%$ of the average local-majority selection gap for Qwen3-8B and Llama-3.1-8B-Instruct, and $63\%$ and $88\%$ of the JP gap. An autopsy shows that more than $95\%$ of these repairs require genuine cross-lingual support rather than a within-language majority, and the bottleneck replicates on a manually audited $483$-problem MATH-500 set. Finally, a controlled training audit shows that rule-GRPO raises trusted accuracy from $0.647$ to $0.735$ while the exact-match reward-error VLB stays high. The unifying message is operational: multilingual RLVR rewards should be audited by language and by answer interface \emph{before} they are optimized.

\end{abstract}

\section{Introduction}

Verifiable reward functions are the central attraction of RLVR. When a task admits an automatic verifier, a model can be optimized against a binary correctness signal without expensive human preference labels. Mathematical reasoning is the canonical case: GSM8K, MATH, and GRPO-style math RL provide widely used problem settings and optimization recipes~\cite{cobbe2021gsm8k,hendrycks2021math,shao2024deepseekmath}. A final-answer verifier maps a generated solution to a reward, and an RL algorithm such as GRPO optimizes the policy against that reward.

This setup carries an implicit reliability assumption: the verifier is a stable proxy for task correctness across the languages and answer formats present in training. In multilingual reasoning, that assumption is fragile. The same numerical answer can surface with currency symbols, markdown emphasis, brackets, decimal variants, full-width digits, CJK magnitude units, or language-specific suffixes. A strict exact-match verifier rejects these semantically correct strings. When such rejections are not distributed uniformly across languages, the verifier converts surface variation into language-dependent reward pressure, and that pressure enters the policy gradient.

We study this phenomenon as \emph{multilingual verifier bias} in RLVR and follow it from measurement to mechanism to consequence. We first build a verifier-robustness suite and a rollout-diagnosis protocol that compares a weak proxy verifier against a trusted canonical-equivalence reward on real model samples. We measure where reward error concentrates by language. We then isolate the mechanism with a controlled plain-numeric interface probe. Finally, we ask what the noise costs at decision time, and find that much of the language gap is not a generation failure but a \emph{selection} failure: the model often already produces the correct answer in some language but selects a wrong one.

Our central empirical object is the language-conditioned reward-error profile and a single scalar summarizing its spread, the verifier language bias $\vlb(m)=\max_l m(l)-\min_l m(l)$ for a per-language metric $m$. The diagnosis is reproducible across three models and two model families, and the worst-affected language is not a property of the benchmark but of the model output distribution.

Taken together, the three measurements are not three separate studies but three views of one quantity---the language-conditioned gap between proxy and trusted reward. The same tuple schema $(\pi,\{x_l\},k,v,r^\ast)$ (Section~\ref{sec:problem}) is the common instrument behind every result in this paper: the rollout diagnosis reads \emph{where} the verifier injects language-conditioned noise directly off the per-language profile, the interface probe isolates \emph{why} (answer format, not arithmetic competence), and the selection bottleneck and training audit then trace \emph{what} that noise costs downstream---at decision time and under optimization. This paper makes three contributions.
\begin{enumerate}
\item \textbf{A general-purpose protocol for auditing multilingual RLVR rewards.} The protocol fixes a single tuple $(\pi,\{x_l\},k,v,r^\ast)$ and a set of language-conditioned reward-error metrics---false-negative rate, false-positive rate, reward-error rate, expected policy pressure, and the verifier language bias $\vlb$---and applies unchanged to any model, verifier, or language set. We release the JP/EN/CN stress suite and three rule-verifier families (exact match, normalized numeric, hybrid) with the paper.
\item \textbf{Reproducible diagnosis with an interface-level mechanism.} On Qwen3-4B, Qwen3-8B, and Llama-3.1-8B-Instruct MGSM rollouts, exact-match reward noise is false-negative dominated and language-conditioned. A same-split plain-numeric probe localizes the exact-match component to the final-answer interface: an interface model drives exact reward-error $\vlb$ to $0.000$ while the residual accuracy gap is unchanged, separating a format effect from a capability effect.
\item \textbf{The cross-lingual selection bottleneck, and a label-free procedure that exploits it.} On fresh MGSM250 rollouts, a simple target-local cross-lingual aggregation procedure---using no trusted labels---recovers $55$--$78\%$ of the average local-majority selection gap and $63$--$88\%$ of the JP gap, replicating on a manually audited $483$-problem MATH-500 set; an autopsy confirms that over $95\%$ of the repairs require genuine cross-lingual agreement, so much of the language gap is a recoverable selection failure, not a generation deficit.
\end{enumerate}

Section~\ref{sec:related} positions the work. Section~\ref{sec:problem} formalizes the metrics. Sections~\ref{sec:suite}--\ref{sec:bottleneck} present the suite, the rollout diagnosis, the interface mechanism, and the selection bottleneck. Section~\ref{sec:training} reports a controlled training audit, and Section~\ref{sec:discussion} draws operational guidance.

\section{Related Work}
\label{sec:related}

\paragraph{Verifier noise in RLVR.}
Verifier noise is increasingly recognized as a first-order concern in RLVR. TinyV studies mathematical false negatives and shows that reducing them improves RL for LLM reasoning~\cite{xu2025tinyv}, and analyses of noisy supervision argue that RLVR is not robust to incorrect reward by default~\cite{zhu2026noisyrlvr}. A theoretical line models verifier unreliability as a stochastic reward channel: the phase-transition statistic $J=\mathrm{TPR}-\mathrm{FPR}$ governs whether incorrect modes are driven out or amplified~\cite{rad2026rateorfate}, and asymmetric false-positive/false-negative rates admit backward and forward corrections that restore an unbiased policy gradient~\cite{cai2025noisyverifier}. Our verifier language bias is the language-conditioned sharpening of exactly this picture: in the false-negative-dominated exact-match regime ($\fpr\approx0$), $J(l)=\mathrm{TPR}(l)-\fpr(l)\approx\mathrm{TPR}(l)=1-\fnr(l)$, so a globally low-noise verifier sits in different reward regimes across languages, and a single global correction cannot track that spread. Reward-hacking studies further motivate a clean separation between a proxy reward and a trusted reward: policies exploit verifier weaknesses, and evaluating that exploitation requires a trusted reference~\cite{helff2026gamingverifiers}. Our work shares this proxy-versus-trusted framing but adds a language axis: we measure how the proxy/trusted gap is distributed across JP/EN/CN and how it enters the policy gradient as language-dependent pressure. Two distinctions matter. First, prior work characterizes verifier reliability largely as a global property of the reward channel; we condition it on language and show that the spread, not the mean, is what biases a multilingual policy. Second, where reward-hacking work studies false positives---rewards granted to wrong answers that the policy then exploits---the multilingual exact-match regime we observe is the opposite, dominated by false negatives, where correct answers are denied reward. The two failure modes call for different audits: false positives are caught by a stronger trusted reference, while the language-conditioned false negatives we study are caught only by conditioning the audit on language and on answer interface.

\paragraph{Multilingual reasoning and judges.}
MGSM established multilingual chain-of-thought math evaluation~\cite{shi2022mgsm}, and cross-lingual collapse studies show that language-centric pretraining shapes reasoning behavior across languages~\cite{park2025crosslingualcollapse}. On the evaluation side, multilingual LLM-as-a-judge work documents that judge reliability varies by language~\cite{fu2025multilingualjudge}. These results concern judges as evaluators; we instead study a rule verifier used as a \emph{reward} and quantify its language-conditioned failure modes during rollouts.

\paragraph{Cross-lingual consistency and aggregation.}
A parallel line improves multilingual reasoning by enforcing agreement across languages. Inference-time frameworks integrate multilingual reasoning paths by majority vote to lift accuracy~\cite{clc2025inference}, and training-time methods enforce cross-lingual self-consistency as an unsupervised RL objective, self-translating each prompt and rewarding cross-language answer agreement~\cite{elhady2026crosslingual}. Our cross-lingual aggregation (Section~\ref{sec:bottleneck}) is neither an accuracy-maximizing decoder nor a training objective: it is a label-free \emph{diagnostic} that selects only among answer clusters already present in the target language, imports no answer from another language (borrowed count $0$), and trains nothing. Its purpose is to measure how much of the residual language gap is a recoverable selection failure rather than a generation deficit---which is evidence for, not a competitor to, the cross-lingual signal these methods exploit, and which our protocol isolates from the format and capability components.

\paragraph{Verifier-free and self-verified rewards.}
A growing line removes or softens the external verifier: NOVER trains without a verifier~\cite{liu2025nover}, VIGOR uses an intrinsic gradient-norm reward~\cite{wen2026vigor}, and Soft-SVeRL uses self-verified soft rewards~\cite{dash2026softsverl}. These are training-time rewards and mark a boundary rather than a competitor to our work. Our cross-lingual aggregation is a label-free, inference-time procedure that recovers much of the selection gap, and our central contribution is the auditing protocol around it.

\paragraph{Math benchmarks across languages.}
Our suite builds on GSM8K and MATH~\cite{cobbe2021gsm8k,hendrycks2021math}, and on the broader argument, made for Portuguese by MATH-PT, that multilingual math evaluation must not rely on English-centric assumptions~\cite{teixeira2026mathpt}. Our focus is complementary: not a new test set, but a protocol for auditing the \emph{reward} that a multilingual RLVR pipeline applies to model outputs.

\section{Problem Formulation}
\label{sec:problem}

Let $x_l$ be a mathematical prompt in language $l \in \{\mathrm{EN}, \mathrm{JP}, \mathrm{CN}\}$, let $y$ be a sampled completion, and let $a(y)$ be its extracted final answer. We distinguish a \emph{trusted} reward $r^\ast(y)$, based on canonical answer equivalence, from a \emph{proxy} reward $r_v(y)$ produced by a verifier $v$ such as exact string match. The trusted reward canonicalizes both prediction and reference (Arabic-digit normalization, CJK magnitude expansion, percentage and sign handling) before comparing, and serves as the reference against which proxy errors are defined.

For a fixed language $l$, a \emph{false negative} (FN) is a completion with $r^\ast=1$ but $r_v=0$ (a correct answer the verifier rejects), and a \emph{false positive} (FP) is a completion with $r^\ast=0$ but $r_v=1$. We report the per-language false-negative and false-positive rates, conditioned on the trusted label,
\begin{align*}
\fnr(l) &= \Pr\nolimits_{y\sim\pi(\cdot\mid x_l)}\!\big[r_v(y)=0 \mid r^\ast(y)=1\big],\\
\fpr(l) &= \Pr\nolimits_{y\sim\pi(\cdot\mid x_l)}\!\big[r_v(y)=1 \mid r^\ast(y)=0\big],
\end{align*}
so that $\fnr(l)$ is the share of trusted-correct completions the verifier rejects. The unconditional reward-error rate is $\errr(l)=\Pr[r_v(y)\neq r^\ast(y)]$, which for an FN-dominated verifier ($\fpr\approx0$) reduces to $\fnr(l)\cdot\mathrm{acc}(l)$, the withheld reward mass. The expected policy pressure $-\errr(l)$ is the signed reward the policy loses per sample on correct-but-rejected answers, and it is the quantity that enters a GRPO-style advantage: a language with a high false-negative rate contributes a systematic negative signal on completions that are in fact correct, biasing the gradient against the formatting habits the model uses in that language.

A diagnosis run is therefore fully specified by a tuple $(\pi, \{x_l\}, k, v, r^\ast)$ and computed by Algorithm~\ref{alg:audit}. Its per-language profile, the $\vlb$ summaries, and the dominant failure mode (FN versus FP) are the protocol's output: the rollout and interface tables instantiate this profile directly, while the selection and training tables report its downstream consequences.

To summarize the language spread of any per-language metric $m$, we define the \textbf{verifier language bias}
\[
\vlb(m) = \max_{l} m(l) - \min_{l} m(l).
\]
The two quantities of primary interest are the reward-error VLB, $\vlb(\errr)$, and the false-negative-rate VLB, $\vlb(\fnr)$. A verifier with $\vlb \approx 0$ treats languages symmetrically even if it is globally noisy; a verifier with large $\vlb$ injects language-specific gradient pressure. In the MGSM setting exact-match errors are FN-dominated ($\fpr \approx 0$), so the false-negative-rate VLB is the most informative single number. Throughout, per-language rates are reported to three decimals, and VLBs, gaps, and language deltas are the differences of those displayed values; confidence intervals are percentile bootstrap.

\begin{figure}[t]
\centering
\begin{tikzpicture}[font=\scriptsize, >=Stealth,
  box/.style={draw, rounded corners, align=center, inner sep=3pt, fill=black!4},
  use/.style={draw, align=center, inner sep=3pt, text width=24mm, minimum height=12mm}]
  \node[use] (u1) at (0,0) {\textbf{Rollout}\\\textbf{diagnosis} (\S\ref{sec:rollout})\\exact-match $v$\\on live rollouts};
  \node[use, right=3mm of u1] (u2) {\textbf{Interface}\\\textbf{probe} (\S\ref{sec:interface})\\control the\\answer format};
  \node[use, right=3mm of u2] (u3) {\textbf{Selection} (\S\ref{sec:bottleneck})\\label-free cross-\\lingual aggregation};
  \node[use, right=3mm of u3] (u4) {\textbf{Training}\\\textbf{audit} (\S\ref{sec:training})\\optimize against $v$};
  \coordinate (mid) at ($(u2.north east)!0.5!(u3.north west)$);
  \node[box, above=9mm of mid] (prof) {Per-language profile: \textsc{fnr}, \textsc{fpr}, \textsc{err}, $\vlb$};
  \node[box, left=6mm of prof] (tuple) {Protocol tuple\\$(\pi,\{x_l\},k,v,r^\ast)$};
  \draw[->] (tuple) -- (prof);
  \foreach \u in {u1,u2,u3,u4} {\draw[->] (prof.south) -- (\u.north);}
\end{tikzpicture}
\caption{One protocol, four uses. A single tuple schema $(\pi,\{x_l\},k,v,r^\ast)$ produces a per-language reward-error profile; varying $\pi$ and $v$ instantiates the four analyses. The metrics are model- and verifier-agnostic, so the protocol transfers unchanged to a new model, verifier, or language set.}
\label{fig:protocol}
\end{figure}
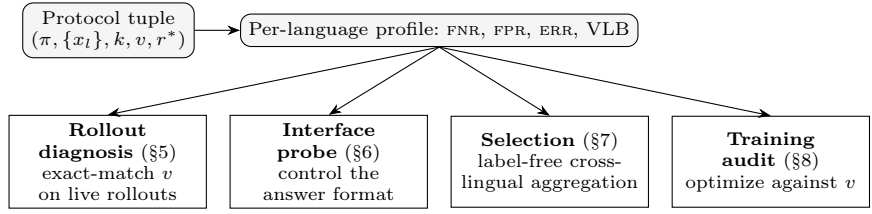

\begin{algorithm}[t]
\caption{Language-Conditioned RLVR Reward Audit}\label{alg:audit}
\begin{algorithmic}[1]
\Require policy $\pi$; multilingual prompts $\{x_l\}_{l\in L}$; sample budget $k$; proxy verifier $v$; trusted reward $r^\ast$
\Ensure per-language profile and $\vlb$ summaries
\For{each language $l\in L$ and prompt $x\in X_l$}
  \State draw $k$ completions $y\sim\pi(\cdot\mid x_l)$; score each under $r_v$ and $r^\ast$
\EndFor
\For{each language $l\in L$}
  \State $\fnr(l)\gets\Pr[r_v{=}0\mid r^\ast{=}1]$;\enspace $\fpr(l)\gets\Pr[r_v{=}1\mid r^\ast{=}0]$;\enspace $\errr(l)\gets\Pr[r_v{\neq}r^\ast]$
\EndFor
\State $\vlb(m)\gets\max_l m(l)-\min_l m(l)$ for $m\in\{\fnr,\fpr,\errr\}$; identify the dominant mode (FN vs.\ FP)
\If{$\fnr(l)>0.3$ and $\vlb(\fnr)>0.2$ for some $l$}
  \State flag $l$; audit the answer interface (\S\ref{sec:interface}), then selection (\S\ref{sec:bottleneck})
\EndIf
\State \Return profile $\{\fnr(l),\fpr(l),\errr(l)\}_l$, $\vlb$ summaries, dominant mode, flagged languages
\end{algorithmic}
\end{algorithm}

\section{A Multilingual Verifier-Robustness Suite}
\label{sec:suite}

The suite starts from public MGSM JP/EN/CN records and available multilingual MATH-500 records. For each seed problem it constructs answer candidates spanning correct answers, format variants, and near misses. The transformations exercise full-width digits, CJK digit and magnitude forms, percentages, explanatory answer text, mixed scripts, and near-miss numeric values. Release scope is the schema, the transformations, the configs, the scripts, and aggregate tables; upstream problem text remains under the original dataset licenses. We treat the suite as a controlled verifier-fragility instrument: a fixed battery on which different rule verifiers can be compared, rather than a sample of organic answer formats.

The trusted reward that anchors every comparison is the canonical-equivalence check of Section~\ref{sec:problem}, which normalizes script, magnitude units, punctuation, and percentage forms on both prediction and reference before deciding equivalence on the normalized values. The proxy verifiers are deliberately weaker so that the gap between proxy and trusted exposes exactly the surface phenomena---script, magnitude unit, punctuation, and wrapper text---that a strict matcher mishandles. This is the design choice that lets a single battery quantify verifier fragility rather than model competence.

We instrument three rule-verifier families: exact string match, normalized numeric equivalence, and a hybrid that tries exact match first and falls back to numeric equivalence. Table~\ref{tab:stress} reports the deterministic cluster rerun. Exact matching is brittle on the synthetic battery, with error around $0.71$ in every language, while normalized numeric and hybrid verifiers reduce error to roughly $0.12$--$0.13$. All three verifiers are FN-dominated on this battery (false-positive rate near zero), and the language spread is small ($\vlb \le 0.013$). The suite provides a controlled fragility baseline; the language-conditioned bias it enables is measured on live rollouts in Section~\ref{sec:rollout}.

\begin{table}[t]
\centering
\caption{Verifier-robustness suite (deterministic cluster rerun). Exact matching is brittle and false-negative dominated; normalized numeric and hybrid verifiers recover most of the loss. The language spread is small, so the suite is a verifier-fragility baseline rather than the language-bias result.}
\label{tab:stress}
\begin{tabularx}{\textwidth}{lCCCC}
\toprule
Verifier & EN err. & JP err. & CN err. & $\vlb(\mathrm{Err})$ \\
\midrule
Exact match & 0.706 & 0.714 & 0.706 & 0.008 \\
Normalized numeric & 0.130 & 0.118 & 0.130 & 0.012 \\
Hybrid exact-then-numeric & 0.130 & 0.118 & 0.130 & 0.012 \\
\bottomrule
\end{tabularx}
\end{table}

\section{Rollout Reward Diagnosis}
\label{sec:rollout}

We now apply the protocol to real rollouts. We sample Qwen3-4B, Qwen3-8B, and Llama-3.1-8B-Instruct on the same MGSM EN/JP/CN split, with $80$ seed problems per language and $k=8$ completions per prompt, giving $1920$ records per model. The trusted reward uses canonical answer equivalence; the primary proxy is an exact-match verifier (\texttt{rule\_exact\_match\_v1}), intentionally weaker than the trusted reward so that format fragility is observed as reward noise. The Llama run provides a second-family replication of the diagnosis.

The suite (Section~\ref{sec:suite}) found exact match brittle but language-symmetric ($\vlb\le0.013$): a fixed battery applies the same format variants to every language and so cannot expose a language-conditioned effect. Live outputs are different---each model develops its own answer-formatting habits, and those habits differ by language---so the reward-noise profile is set by the interaction between the verifier's brittleness and the model's language-conditioned formatting, which only a live policy reveals. The language symmetry of Table~\ref{tab:stress} breaks sharply here.

\begin{figure}[t]
\centering
\includegraphics[width=0.78\textwidth]{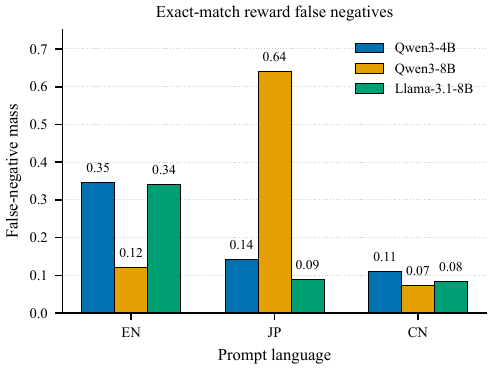}
\caption{False-negative rate by language on MGSM $k=8$ rollouts. The worst-affected language shifts across models: EN for Qwen3-4B and Llama-3.1-8B-Instruct, JP for Qwen3-8B.}
\label{fig:fnmass}
\end{figure}

Table~\ref{tab:rollout} reports the per-language profile, and Fig.~\ref{fig:fnmass} visualizes the false-negative rate. Two facts stand out. First, the reward noise is essentially all false-negative: the observed false-positive rate is $0$ for all three models in every language, so the verifier never rewards a wrong answer here---it only rejects correct ones. Second, the false-negative rate is strongly language-conditioned, and the worst language depends on the model. Among trusted-correct completions, Qwen3-4B is rejected most on EN ($\fnr=0.345$), Qwen3-8B on JP ($\fnr=0.642$, against $0.122$ on EN and $0.073$ on CN), and Llama-3.1-8B-Instruct again on EN ($0.342$). The corresponding false-negative-rate VLB ranges from $0.234$ (Qwen3-4B) through $0.258$ (Llama) to $0.569$ (Qwen3-8B). A problem-level bootstrap ($B=4000$, resampling the $80$ shared problems) confirms these are not sampling artifacts: the VLB $95\%$ confidence intervals are $[0.16,0.32]$, $[0.46,0.67]$, and $[0.17,0.37]$ respectively, all excluding zero, and the Qwen3-8B JP rate of $0.642$ has CI $[0.55,0.73]$.

The Qwen3-8B JP figure is the sharpest case: $64.2\%$ of its trusted-correct JP completions are rejected by exact match, so an RLVR loop driven by exact match would systematically suppress correct JP reasoning. These rejections have a single dominant cause rather than diffuse arithmetic disagreement: $307$ of the $308$ rejected-correct JP completions ($99.7\%$) wrap the answer in angle brackets (e.g.\ \texttt{<18>} for $18$), and $94.8\%$ recover the gold answer after stripping non-numeric characters---so a numeric-equivalence fallback alone would readmit nearly all of them. Each model's worst-affected language is dominated by one wrapper format: markdown or math delimiters for Qwen3-4B (EN, $98\%$ of its FN) and Llama-3.1-8B-Instruct (EN, $93\%$), and angle brackets for Qwen3-8B (JP, $99.7\%$); in every case numeric normalization readmits most of the rejected-correct mass ($83\%$, $70\%$, and $95\%$ respectively). Across models, the verifier bias is one formatting habit per model, not a reasoning gap. An aggregate verifier that looks acceptable on a static battery (Table~\ref{tab:stress}) can still inject large, model-specific language pressure once it meets a live policy.

Trusted accuracy and the false-negative rate are different axes. Qwen3-8B has its \emph{highest} trusted accuracy on EN ($0.948$) and a respectable $0.750$ on JP, yet its reward noise is overwhelmingly concentrated on JP. The model is largely able to solve the JP problems; it is the exact-match verifier that fails to credit the JP answers. This is the regime in which verifier bias is most dangerous, because the loss is invisible to a trusted-accuracy report and only appears when the proxy reward is inspected by language. Conversely, Llama-3.1-8B-Instruct has low trusted accuracy on JP ($0.469$) but a \emph{low} JP false-negative rate ($0.090$): when the model rarely produces a correct JP answer in the first place, the reward error migrates to EN, where the model is both productive and idiosyncratic in its formatting. The diagnosis thus distinguishes a capability-limited language from a verifier-penalized one, a distinction a single accuracy number cannot make.

Because the observed false-positive rate is zero, the entire reward error here is withheld deserved reward, never granted undeserved reward. This is the more corrosive direction for RLVR: false negatives directly depress the advantage of correct trajectories and, concentrated in one language, inject a systematic language-specific negative signal into the policy gradient. The remaining sections trace where this withheld reward comes from (Section~\ref{sec:interface}) and what it costs at decision time (Section~\ref{sec:bottleneck}).

\begin{table}[t]
\centering
\caption{Rollout reward diagnosis on MGSM EN/JP/CN, $80$ seed problems per language, $k=8$ ($1920$ records per model). FN is the false-negative rate among trusted-correct completions. Exact-match reward noise is false-negative dominated (observed $\fpr=0$ throughout) and language-conditioned, and the worst language is model-distribution dependent.}
\label{tab:rollout}
\begin{tabularx}{\textwidth}{lCCCCCCC}
\toprule
\multirow{2}{*}{Model} & \multicolumn{3}{c}{Trusted acc.} & \multicolumn{3}{c}{FN rate} & \multirow{2}{*}{$\vlb(\fnr)$} \\
\cmidrule(lr){2-4}\cmidrule(lr){5-7}
 & EN & JP & CN & EN & JP & CN & \\
\midrule
Qwen3-4B & 0.933 & 0.708 & 0.819 & 0.345 & 0.143 & 0.111 & 0.234 \\
Qwen3-8B & 0.948 & 0.750 & 0.813 & 0.122 & 0.642 & 0.073 & 0.569 \\
Llama-3.1-8B-Instruct & 0.817 & 0.469 & 0.539 & 0.342 & 0.090 & 0.084 & 0.258 \\
\bottomrule
\end{tabularx}
\end{table}

\section{The Interface Mechanism}
\label{sec:interface}

The rollout diagnosis shows \emph{that} exact-match reward noise is language-conditioned, but not \emph{why}. We test two hypotheses with a controlled probe: does the noise reflect genuine arithmetic differences across languages, or a surface mismatch at the final-answer interface?

We construct a leak-free MGSM heldout split of $20$ problem identifiers, disjoint from any training prompts, and evaluate three Qwen3-8B variants under a \emph{plain-numeric} prompt that requests a bare numeric final answer: the base model, a same-split rule-GRPO model, and a final-answer-interface SFT model that is supervised only on producing the answer in the plain-numeric form. Each variant is evaluated on $240$ records (the $20$ identifiers in three languages with $k=4$) with no rollout failures, under both the exact-match and normalized-numeric diagnoses. Table~\ref{tab:interface} reports the result.

Two effects separate cleanly. Under the plain-numeric interface the exact-match noise collapses: the base model's exact reward-error VLB is already low at $0.013$, and the interface SFT model drives both it and the exact false-negative-rate VLB to $0.000$; a problem-level bootstrap puts the interface model's exact VLB at $[0,0]$ (Base $[0,0.05]$, rule-GRPO $[0,0.14]$), so the format component of the bias is robustly eliminated. At the same time the language accuracy gap does not move with the interface fix: it is $0.150$ for both rule-GRPO and the interface model, against $0.213$ for the base model. In other words, controlling the answer interface removes the measured exact-match verifier bias but leaves a residual accuracy difference between languages---a policy capability difference, not a verifier-format artifact---whose scale and composition Section~\ref{sec:bottleneck} establishes on fresh data at the $483$-problem scale.

The per-language trusted accuracies of the interface model make the residual concrete: $0.875$ on EN, $0.825$ on CN, and $0.725$ on JP, so the $0.150$ gap is the EN--JP difference that survives once formatting is held constant. The contrast with the rollout diagnosis is informative. In Section~\ref{sec:rollout} the dominant Qwen3-8B problem was a JP \emph{verifier} penalty; here, under a controlled plain-numeric interface, the JP verifier penalty is gone (exact $\vlb(\fnr)=0$) and what remains is a smaller JP \emph{accuracy} deficit. The probe therefore does two things at once: it confirms that the large exact-match component of the bias was a format artifact, and it isolates the genuine, much smaller, trusted-accuracy gap that any reward design must still contend with. The same split, the same model, and the same diagnosis pipeline are held fixed while only the answer interface varies, so the decomposition into format artifact and capability gap is direct.

\begin{table}[t]
\centering
\caption{Plain-numeric, same-split interface probe on Qwen3-8B ($240$ records per variant, zero rollout failures). Controlling the final-answer interface drives the exact-match verifier bias to zero, while the residual language accuracy gap is unchanged.}
\label{tab:interface}
\begin{tabularx}{\textwidth}{lCCCC}
\toprule
Variant & Avg.\ trusted acc. & Lang.\ gap & Exact $\vlb(\errr)$ & Exact $\vlb(\fnr)$ \\
\midrule
Base & 0.804 & 0.213 & 0.013 & 0.018 \\
Rule-GRPO & 0.817 & 0.150 & 0.050 & 0.069 \\
Interface-SFT & 0.808 & 0.150 & 0.000 & 0.000 \\
\bottomrule
\end{tabularx}
\end{table}

The exact-match component of multilingual verifier bias is an interface problem: it is removed by fixing how the final answer is written and parsed, not by changing the RL objective. What remains after the interface is controlled is a residual trusted-accuracy gap between languages, and Section~\ref{sec:bottleneck} shows that a large share of \emph{that} gap is itself recoverable at selection time.

\section{The Cross-Lingual Selection Bottleneck}
\label{sec:bottleneck}

If the residual gap in Section~\ref{sec:interface} is a capability gap, the next question is whether it is a \emph{generation} gap (the model never produces the correct answer in the weak language) or a \emph{selection} gap (the model produces it among its $k$ samples but a majority vote picks a wrong one). We answer this on fresh data.

\paragraph{Setup.}
We sample Qwen3-8B and Llama-3.1-8B-Instruct on a fresh MGSM250 split ($250$ problems per language, disjoint from the diagnosis split) with $k=8$. The baseline decision rule is \emph{local} majority: for each problem in language $l$, take the most frequent canonicalized answer among the $k$ same-language samples. The comparison rule is \emph{target-local cross-lingual aggregation}: the selector may consult same-problem answer-cluster counts from the other two languages, but it chooses only among clusters that actually appear in the target language, never uses trusted labels, and runs entirely at inference time.

\begin{figure}[t]
\centering
\includegraphics[width=\textwidth]{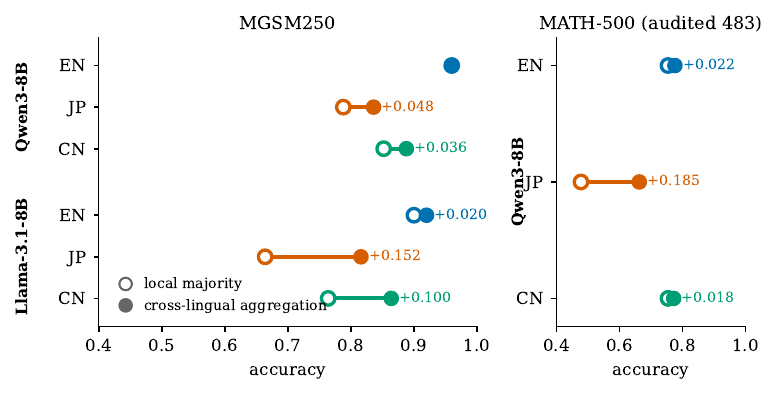}
\caption{Cross-lingual selection bottleneck: per-language accuracy under local majority (open) versus target-local cross-lingual aggregation (filled). Left: MGSM250 ($250$ problems/language; Qwen3-8B and Llama-3.1-8B-Instruct). Right: the audited $483$-problem MATH-500 set (Qwen3-8B). The procedure uses no trusted labels and imports no cross-language answer; the JP gain dominates.}
\label{fig:selection}
\end{figure}

\paragraph{Result.}
Table~\ref{tab:bottleneck} reports the outcome and Fig.~\ref{fig:selection} visualizes it. For both models, aggregation lifts average accuracy and shrinks the language gap, and the gains concentrate on JP. For Qwen3-8B, average accuracy rises from $0.867$ to $0.895$ and the EN--JP gap falls from $0.172$ to $0.124$; for Llama-3.1-8B-Instruct, average accuracy rises from $0.776$ to $0.867$ and the gap falls from $0.236$ to $0.104$. Measured against the pass@$k$ ceiling, aggregation closes $55.3\%$ and $78.2\%$ of the average local-majority selection gap, and $63.2\%$ and $88.4\%$ of the JP gap. The effect is statistically strong: of the available selection opportunities, Qwen3-8B fixes $24$ of $38$ and Llama $71$ of $87$, against $3$ regressions each (paired sign-test $p=4.9\times10^{-5}$ for Qwen3-8B and $p<10^{-12}$ for Llama). Among the decisions aggregation changes, the fraction corrected---$24/27$ and $71/74$---is $0.889$ and $0.959$.

\begin{table}[t]
\centering
\caption{Fresh MGSM250 selection bottleneck ($k=8$, $250$ problems per language). \emph{Local} is same-language majority; \emph{X-ling.} is target-local cross-lingual aggregation, which uses no trusted labels and only selects among clusters present in the target language. ``Gap closed'' is the fraction of the local-to-pass@$k$ selection gap recovered.}
\label{tab:bottleneck}
\setlength{\tabcolsep}{4.5pt}
\begin{tabularx}{\textwidth}{lCCCCCCCC}
\toprule
\multirow{2}{*}{Model} & \multicolumn{2}{c}{Avg.\ acc.} & \multicolumn{2}{c}{EN--JP gap} & \multicolumn{2}{c}{Gap closed} & \multirow{2}{*}{Fixed} & \multirow{2}{*}{Reg.} \\
\cmidrule(lr){2-3}\cmidrule(lr){4-5}\cmidrule(lr){6-7}
 & Local & X-ling. & Local & X-ling. & Avg. & JP & & \\
\midrule
Qwen3-8B & 0.867 & 0.895 & 0.172 & 0.124 & 55.3\% & 63.2\% & 24/38 & 3 \\
Llama-3.1-8B-Instruct & 0.776 & 0.867 & 0.236 & 0.104 & 78.2\% & 88.4\% & 71/87 & 3 \\
\bottomrule
\end{tabularx}
\end{table}

\paragraph{Autopsy: the repairs are genuinely cross-lingual.}
Could these repairs be $k=8$ sampling noise---the correct cluster already the within-language plurality, only confirmed by aggregation? We tested every fixed case by comparing the target-language vote for the selected cluster against its sibling support, and the data rule it out: for Qwen3-8B, $23$ of $24$ fixes ($95.8\%$) have a target-language vote no greater than sibling support (Llama: $70$ of $71$, $98.6\%$), and both siblings support the chosen cluster in $22/24$ Qwen and $60/71$ Llama fixes. Because the selector chooses only among clusters present in the target language, no answer is ever imported (borrowed count $0$). The repairs are cross-lingual selection signal, not within-language sampling artifacts.

\paragraph{Robustness.}
The aggregation rule weights sibling support by a single exponent $\alpha$; the result is insensitive to it, with average accuracy changing by $\le0.004$ across $\alpha\in\{0.5,1.0,2.0\}$ and at least $91.7\%$ of fixes preserved (we report $\alpha=1.0$). A conservative variant that triggers aggregation only when both sibling languages support a target cluster retains $22$ of $24$ Qwen3-8B fixes while reducing regressions from $3$ to $2$ at the same gap reduction, so sibling agreement is itself a usable reliability gate.

\paragraph{Replication on a harder benchmark.}
The same target-local rule replicates on a harder external set: the full MATH-500 in EN/JP/CN.\footnote{We use the public multilingual release \texttt{appier-ai-research/Multilingual-MATH-500}; gold answers are unchanged. The Japanese and Chinese problems are machine translations; we manually audit the Japanese problems, to which the exclusion below applies.} By a criterion independent of model outputs---Japanese answerability---two reviewers audited all $500$ problems and excluded $17$ whose Japanese rendering had dropped a figure-defining Asymptote block or contained answer-altering translation artifacts, leaving $483$. On this audited set, Qwen3-8B local-majority average accuracy is $0.662$ (EN $0.754$, JP $0.478$, CN $0.754$; EN--JP gap $0.276$), and aggregation raises it to $0.737$ (EN $0.776$, JP $0.663$, CN $0.772$; gap $0.113$), with $123$ corrections against $14$ regressions among the $137$ changed selection decisions (paired sign-test $p<10^{-20}$). The JP gain ($+0.185$) again dominates and borrowing stays at $0$, so the selection bottleneck holds at scale on a harder benchmark, not only on MGSM.

\paragraph{Interpretation.}
A large fraction of the residual language gap is not the model failing to reason in the weak language; it is the model failing to \emph{select} a correct answer it has already produced, which sibling-language agreement identifies without any trusted label. This completes a three-way decomposition of the language-conditioned reward gap: a \emph{format artifact} (Section~\ref{sec:interface}, removable at the answer interface), a \emph{selection failure} (this section, recoverable by a label-free cross-lingual signal), and a residual \emph{generation deficit} (the part of the pass@$k$ ceiling that neither addresses). The diagnosis thus locates recoverable headroom and points future verifier-free work toward selection and reranking rather than only toward generation quality.

\section{Training-Time Consequence}
\label{sec:training}

Once the verifier is actually used as an RLVR reward, the diagnosis and its interface mechanism carry direct consequences. Two predictions follow from the preceding sections, and a controlled rule-GRPO audit confirms both.

\paragraph{Reward variance (a consequence of Section~\ref{sec:rollout}).}
The diagnosis predicts a failure of the learning signal itself. A Qwen3-4B rule-GRPO run on a single-language split has zero reward standard deviation across all five logged steps---every sampled group receives identical reward, so there is no gradient (gradient norm $0.000$). This is exactly the FN-dominated regime of Section~\ref{sec:rollout}: when correct answers are uniformly accepted or uniformly rejected within a group, the group reward is constant. Diagnosis-guided mixed-group sampling, which deliberately mixes languages and difficulty within a group, restores a nonzero signal for Qwen3-8B---mean reward standard deviation $0.258$, nonzero variance in $5$ of $10$ steps, and gradient norm $0.172$.

\paragraph{Persisting bias (a consequence of Section~\ref{sec:interface}).}
The interface mechanism predicts that optimizing the proxy reward raises capability without touching the format-induced bias. On a held $17$-prompt mixed split with $k=4$ ($68$ records per model, record-weighted across languages), trusted accuracy improves from $0.647$ to $0.735$, with gains on CN, EN, and JP, yet the exact-match reward-error VLB worsens from $0.417$ to $0.500$ (Table~\ref{tab:rulegrpo}). Capability and verifier bias are orthogonal: accuracy improvement does not imply bias reduction, and the selection headroom identified in Section~\ref{sec:bottleneck} is not recovered by proxy optimization. This is the operational argument for measuring the bias as its own quantity \emph{before} optimizing against it.

\begin{table}[t]
\centering
\caption{Rule-GRPO adapter evaluation ($17$-prompt mixed split, $k=4$, $68$ records per model). Trusted accuracy improves while the exact-match reward-error VLB stays high.}
\label{tab:rulegrpo}
\begin{tabularx}{\textwidth}{lCCCCC}
\toprule
Variant & CN acc. & EN acc. & JP acc. & Avg.\ acc. & $\vlb(\mathrm{Err})$ \\
\midrule
Base & 0.417 & 0.800 & 0.750 & 0.647 & 0.417 \\
Rule-GRPO & 0.458 & 0.900 & 0.875 & 0.735 & 0.500 \\
\bottomrule
\end{tabularx}
\end{table}

\section{Discussion}
\label{sec:discussion}

\paragraph{The workflow.}
The results compose into one workflow. The rollout diagnosis (Section~\ref{sec:rollout}) locates where a verifier injects language-conditioned reward error; the interface probe (Section~\ref{sec:interface}) splits that error into a fixable answer-format problem and a residual capability gap; and the selection bottleneck (Section~\ref{sec:bottleneck}) shows that much of the residual gap is recoverable headroom. The training audit (Section~\ref{sec:training}) shows that skipping these steps is costly: a verifier can silently zero out the learning signal, and optimizing against it raises accuracy while leaving the language bias untouched.

\paragraph{An operational rule of thumb.}
The practitioner trigger in Algorithm~\ref{alg:audit}---a single language with false-negative rate above $\sim0.3$ and false-negative-rate VLB above $\sim0.2$---emerges from the three models, and the ordering it prescribes (interface first, where the exact-match component lives; then selection, where sibling-language agreement recovers correct answers without labels) held across all three here.

\paragraph{Implications for verifier-free rewards.}
The selection bottleneck has a direct bearing on the verifier-free line of work. Methods that remove the external verifier---training without one~\cite{liu2025nover}, using an intrinsic gradient signal~\cite{wen2026vigor}, or self-verifying with soft rewards~\cite{dash2026softsverl}---implicitly hope that an internal signal can stand in for trusted correctness. Our diagnosis suggests where such a signal is most needed and most attainable: not in generating correct answers in the weak language, which the models already do at the pass@$k$ level, but in \emph{selecting} among answers the model has produced. The cross-lingual aggregation result is evidence that a label-free signal with this property exists: sibling-language agreement identifies the correct cluster in over $95\%$ of the repaired cases.

\paragraph{Reusability.}
The same tuple $(\pi, \{x_l\}, k, v, r^\ast)$ and the same model- and verifier-agnostic metrics---false-negative rate, false-positive rate, reward-error rate, and $\vlb$---drive every table here, from a synthetic battery to three live policies, a controlled interface probe, and a training run, read on a common scale. This uniformity is the protocol's contribution: it turns ``is my multilingual verifier fair?'' into a measurement with a fixed output format.

\section{Conclusion}

We presented a reusable measurement loop for multilingual verifier bias in RLVR and used it to establish three results. Exact-match verifier reward creates false-negative-dominated, language-conditioned pressure on real Qwen and Llama MGSM rollouts, with the worst language set by the model rather than the benchmark. A controlled plain-numeric probe localizes the exact-match component to the final-answer interface, separating a removable format effect from a residual capability gap. On fresh data, a label-free cross-lingual selection rule closes $55$--$78\%$ of the average local-majority selection gap and $63$--$88\%$ of the JP gap, with an autopsy confirming that the repairs are genuinely cross-lingual. Together these results argue for a concrete discipline: audit multilingual RLVR rewards by language and by answer interface before optimizing against them. The diagnosis spans two model families on MGSM JP/EN/CN and is corroborated on a manually audited $483$-problem MATH-500 set; the cross-lingual aggregation runs at inference time on parallel multilingual samples, and turning its selection signal into a training-time reward is a natural next step.

\bibliographystyle{splncs04}
\bibliography{references}

@misc{shao2024deepseekmath,
  title = {DeepSeekMath: Pushing the Limits of Mathematical Reasoning in Open Language Models},
  author = {Shao, Zhihong and Wang, Peiyi and Zhu, Qihao and Xu, Runxin and Song, Junxiao and Bi, Xiao and Zhang, Haowei and Zhang, Mingchuan and Li, Y. K. and Wu, Y. and Guo, Daya},
  year = {2024},
  eprint = {2402.03300},
  archivePrefix = {arXiv},
  url = {https://arxiv.org/abs/2402.03300}
}

@misc{liu2025nover,
  title = {NOVER: Incentive Training for Language Models via Verifier-Free Reinforcement Learning},
  author = {Liu, Wei and Qi, Siya and Wang, Xinyu and Qian, Chen and Du, Yali and He, Yulan},
  year = {2025},
  eprint = {2505.16022},
  archivePrefix = {arXiv},
  url = {https://arxiv.org/abs/2505.16022}
}

@misc{wen2026vigor,
  title = {Verifier-Free RL for LLMs via Intrinsic Gradient-Norm Reward},
  author = {Wen, Xuexiang and Yu, Hang and Zhu, Linchao and Wang, Gaoang},
  year = {2026},
  eprint = {2605.09920},
  archivePrefix = {arXiv},
  url = {https://arxiv.org/abs/2605.09920}
}

@misc{dash2026softsverl,
  title = {Soft-SVeRL: Self-Verified Reinforcement Learning with Soft Rewards},
  author = {Dash, Saurabh and Clavier, Pierre and Dang, John and Galle, Matthias and Fadaee, Marzieh and {\"U}st{\"u}n, Ahmet and Ermis, Beyza},
  year = {2026},
  eprint = {2605.28561},
  archivePrefix = {arXiv},
  url = {https://arxiv.org/abs/2605.28561}
}

@misc{park2025crosslingualcollapse,
  title = {Cross-lingual Collapse: How Language-Centric Foundation Models Shape Reasoning in Large Language Models},
  author = {Park, Cheonbok and Kim, Jeonghoon and Lee, Joosung and Bae, Sanghwan and Choo, Jaegul and Yoo, Kangmin},
  year = {2025},
  eprint = {2506.05850},
  archivePrefix = {arXiv},
  url = {https://arxiv.org/abs/2506.05850}
}

@misc{shi2022mgsm,
  title = {Language Models are Multilingual Chain-of-Thought Reasoners},
  author = {Shi, Freda and Suzgun, Mirac and Freitag, Markus and Wang, Xuezhi and Srivats, Suraj and Vosoughi, Soroush and Chung, Hyung Won and Tay, Yi and Ruder, Sebastian and Zhou, Denny and Das, Dipanjan and Wei, Jason},
  year = {2022},
  eprint = {2210.03057},
  archivePrefix = {arXiv},
  url = {https://arxiv.org/abs/2210.03057}
}

@misc{xu2025tinyv,
  title = {TinyV: Reducing False Negatives in Verification Improves RL for LLM Reasoning},
  author = {Xu, Zhangchen and Li, Yuetai and Jiang, Fengqing and Ramasubramanian, Bhaskar and Niu, Luyao and Lin, Bill Yuchen and Poovendran, Radha},
  year = {2025},
  eprint = {2505.14625},
  archivePrefix = {arXiv},
  url = {https://arxiv.org/abs/2505.14625}
}

@misc{zhu2026noisyrlvr,
  title = {Noisy Data is Destructive to Reinforcement Learning with Verifiable Rewards},
  author = {Zhu, Yuxuan and Kang, Daniel},
  year = {2026},
  eprint = {2603.16140},
  archivePrefix = {arXiv},
  url = {https://arxiv.org/abs/2603.16140}
}

@misc{fu2025multilingualjudge,
  title = {How Reliable is Multilingual LLM-as-a-Judge?},
  author = {Fu, Xiyan and Liu, Wei},
  year = {2025},
  eprint = {2505.12201},
  archivePrefix = {arXiv},
  url = {https://arxiv.org/abs/2505.12201}
}

@misc{helff2026gamingverifiers,
  title = {LLMs Gaming Verifiers: RLVR can Lead to Reward Hacking},
  author = {Helff, Lukas and Delfosse, Quentin and Steinmann, David and Harle, Ruben and Shindo, Hikaru and Schramowski, Patrick and Stammer, Wolfgang and Kersting, Kristian and Friedrich, Felix},
  year = {2026},
  eprint = {2604.15149},
  archivePrefix = {arXiv},
  url = {https://arxiv.org/abs/2604.15149}
}

@misc{cobbe2021gsm8k,
  title = {Training Verifiers to Solve Math Word Problems},
  author = {Cobbe, Karl and Kosaraju, Vineet and Bavarian, Mohammad and Chen, Mark and Jun, Heewoo and Kaiser, Lukasz and Plappert, Matthias and Tworek, Jerry and Hilton, Jacob and Nakano, Reiichiro and Hesse, Christopher and Schulman, John},
  year = {2021},
  eprint = {2110.14168},
  archivePrefix = {arXiv},
  url = {https://arxiv.org/abs/2110.14168}
}

@misc{hendrycks2021math,
  title = {Measuring Mathematical Problem Solving With the MATH Dataset},
  author = {Hendrycks, Dan and Burns, Collin and Kadavath, Saurav and Arora, Akul and Basart, Steven and Tang, Eric and Song, Dawn and Steinhardt, Jacob},
  year = {2021},
  eprint = {2103.03874},
  archivePrefix = {arXiv},
  url = {https://arxiv.org/abs/2103.03874}
}

@misc{teixeira2026mathpt,
  title = {MATH-PT: A Math Reasoning Benchmark for European and Brazilian Portuguese},
  author = {Teixeira, Tiago and Erthal, Ana Carolina and Belieni, Juan and Canaverde, Beatriz and Mesquita, Diego and Faria, Miguel and da Silva, Eliezer de Souza and Martins, Andre F. T.},
  year = {2026},
  eprint = {2604.25926},
  archivePrefix = {arXiv},
  url = {https://arxiv.org/abs/2604.25926}
}

@misc{cai2025noisyverifier,
  title = {Reinforcement Learning with Verifiable yet Noisy Rewards under Imperfect Verifiers},
  author = {Cai, Xin-Qiang and Wang, Wei and Liu, Feng and Liu, Tongliang and Niu, Gang and Sugiyama, Masashi},
  year = {2025},
  eprint = {2510.00915},
  archivePrefix = {arXiv},
  url = {https://arxiv.org/abs/2510.00915}
}

@misc{rad2026rateorfate,
  title = {Rate or Fate? {RLV}$^\varepsilon${R}: Reinforcement Learning with Verifiable Noisy Rewards},
  author = {Rad, Ali and Filom, Khashayar and Keivan, Darioush and {Mohajerin Esfahani}, Peyman and Kamalinejad, Ehsan},
  year = {2026},
  eprint = {2601.04411},
  archivePrefix = {arXiv},
  url = {https://arxiv.org/abs/2601.04411}
}

@misc{clc2025inference,
  title = {Cross-Lingual Consistency: A Novel Inference Framework for Advancing Reasoning in Large Language Models},
  author = {Yu, Zhiwei and Li, Tuo and Wang, Changhong and Chen, Hui and Zhou, Lang},
  year = {2025},
  eprint = {2504.01857},
  archivePrefix = {arXiv},
  url = {https://arxiv.org/abs/2504.01857}
}

@misc{elhady2026crosslingual,
  title = {Cross-lingual Self-Consistency for Multilingual Reasoning with Language Models},
  author = {Elhady, Ahmed and Agirre, Eneko and Artetxe, Mikel},
  year = {2026},
  eprint = {2606.01464},
  archivePrefix = {arXiv},
  url = {https://arxiv.org/abs/2606.01464}
}

\end{document}